\documentclass[times,twocolumn,final]{elsarticle} 

\usepackage{prletters}
\usepackage{framed,multirow}
\usepackage{makecell}

\usepackage{amssymb}
\usepackage{latexsym}
\usepackage{amsmath}
\usepackage{amsfonts}
\usepackage{amstext}
\usepackage{amsthm}
\usepackage{amsmath}
\usepackage{subcaption}
\usepackage{comment}
\usepackage[ruled,vlined]{algorithm2e}
\usepackage{booktabs}
\usepackage{dsfont}
\usepackage[inline]{enumitem}

\usepackage{url}
\usepackage[table,xcdraw]{xcolor}
\definecolor{newcolor}{rgb}{.8,.349,.1}
\definecolor{ao(english)}{rgb}{0.0, 0.5, 0.0}
\usepackage{soul}

\journal{Pattern Recognition Letters}

\begin{document}
	\thispagestyle{empty}

	\ifpreprint
	\setcounter{page}{1}
	\else
	\setcounter{page}{1}
	\fi
	
	\begin{frontmatter}
		
		\title{
  A Siamese-based Verification System for Open-set Architecture Attribution of Synthetic Images
 }
		
		\author[1]{Lydia \snm{Abady}\corref{cor1}}
		\cortext[cor1]{Corresponding author}
		\ead{lydia.abady@unisi.it}
		\author[1]{Jun \snm{Wang}}
		\ead{j.wang@student.unisi.it}
		\author[1]{Benedetta \snm{Tondi}}
		\ead{benedetta.tondi@unisi.it}
		\author[1]{Mauro \snm{Barni}}
		\ead{barni@dii.unisi.it}

		\address[1]{Department of Information Engineering and Mathematics, University of Siena, Via Roma 56, 53100 Siena, Italy}

		\begin{abstract}
			Despite the wide variety of methods developed for synthetic image attribution, most of them can only attribute images generated by models or architectures included in the training set and do not work with {\em unknown} architectures, hindering their applicability in real-world scenarios. In this paper, we propose a verification framework that relies on a Siamese Network to address the problem of open-set attribution of synthetic images to the architecture that generated them. We consider two different settings. In the first setting, the system determines whether two images have been produced by the same generative architecture or not. In the second setting, the system verifies a claim about the architecture used to generate a synthetic image, utilizing one or multiple reference images generated by the claimed architecture. The main strength of the proposed system is its ability to operate in both closed and open-set scenarios so that the input images, either the query and reference images, can belong to the architectures considered during training or not. Experimental evaluations encompassing various generative architectures such as GANs, diffusion models, and transformers, focusing on synthetic face image generation, confirm the excellent performance of our method in both closed and open-set settings, as well as its strong generalization capabilities.
   		\end{abstract}
		
		
	\end{frontmatter}

	\section{Introduction}
	\label{sec.intro}

Synthetic manipulation and generation of images have become ubiquitous and are being increasingly used in a wide variety of applications. Contents generated by Artificial Intelligence (AI) and deepfake technology have garnered widespread attention because they are often used with malicious intent, thus representing a serious threat to public trust. In response to this, several methods have been developed for the detection of synthetic images, performing real vs fake classification. However, in many cases, only knowing that the image is fake is not enough and more information is required on the synthetic nature of the image.
In particular, in some cases, it is necessary to know the specific model or the type of architecture used to produce the fake image (synthetic image attribution). Several methods have been proposed for model-level attribution via multi-class classifiers by relying on the artifacts or signatures (fingerprints) left by the models in the images they generate \cite{marra2019gans,yuresponsible,yu2021artificial}.
With model-level attribution, models that are fine-tuned or retrained with a different configuration, for instance by using a different initialization or training data, are considered different models, as they are characterized by different fingerprints.
This can be a limitation in many real-world applications, where model-level granularity is not needed or is too difficult to achieve. As an answer, recent approaches have started addressing the attribution task under a more general setting, attributing the synthetic images to the architecture that was used to generate them, instead of the specific model  \cite{yang2022deepfake,bui2022repmix}.

A common drawback of most model-level and architecture-level attribution methods \cite{yang2022deepfake,bui2022repmix} is that they can not work in an open-set scenario wherein the test images are generated by a model/architecture that has not been considered during training. This seriously limits the applicability of these methods in real-world applications, where the images seen at operation time may be produced by models/architectures that have not been seen during training, with the consequence that the predictions made by the methods are not trustable.
Some approaches address this issue by performing classification with a rejection class, revealing unknown models/architectures, and refraining from identifying the model/architecture, in this case, \cite{girish2021towards,yang2023progressive,wang2023open}.

In this paper, we adopt a different approach, treating the synthetic architecture attribution task as a verification task. 
In particular, we propose a method to decide whether two synthetic input images have been produced by the same generative architecture or not.
\footnote{Focusing on attribution of synthetic images, our system do not consider pristine data as input.}
%
We also consider a slightly different setting, where the system is asked to verify a claim about the architecture used to generate a given  image, by relying on multiple reference images produced by the same architecture.
The system is based on a Siamese Network architecture with an EfficientNet-B4 backbone, trained in two phases: the first one focusing on the feature extraction part to learn the embeddings and the second on the final decision layers.
We carried out a thorough experimental campaign considering several generative architectures, including Generative Adversarial Networks (GANs), diffusion models, and transformers, by focusing on synthetic face image generation. The results showed that the proposed system performs very well in both closed and open-set settings, with a significant advantage with respect to systems based on the introduction of a rejection class, which are not able to provide any information about out-of-set architectures, other than recognizing that they do not belong to the set used for training.

The  contributions of this paper can be summarized as follows:
	\begin{itemize}
		\item We propose a new verification framework for open-set architecture attribution of synthetic images. Two verification scenarios are considered: in the first one, the system has to decide whether two synthetic input images are generated by the same architecture or not; in the second one, an synthetic image and a claim on the architecture generating it are given, and the system has to decide whether to support the claim or not.
  \item By focusing on the face image generation domain, we run extensive experiments that prove the good performance of the proposed verification method with several types of generative architectures when different combinations of architectures are considered in both closed and open set scenarios.
		\item We perform several generalization tests proving that the system can verify the architecture also when unknown models for the various architectures are considered to produce the test images, e.g. models trained with different pristine data, different training procedures, and different configurations of parameters.

		\item
  We exploit the verification architecture to build a system for classification with a rejection option, showing that the proposed system outperforms state-of-the-art methods for open-set architecture attribution with a rejection class.
	\end{itemize}
	
	The rest of the paper is organized as follows: in Section \ref{sec.sota}, we briefly review the state-of-the-art of synthetic image generation and attribution.
Then, in Section \ref{sec.arch}, we describe the proposed framework and architecture. In Section \ref{sec.method}, we describe the datasets and the methodology, including the training procedure and the verification protocol. The results of the experiments we carried out to validate the effectiveness of the proposed verification system are reported in Section \ref{sec.results}.

	\section{Related Works}
	\label{sec.sota}

To build our 
verification system for synthetic image attribution,  we resort to a Siamese Network-based architecture. It is proper to stress that the use of  Siamese Networks  is not novel in the  forensic literature, and several approaches have been proposed that relies on contrastive learning, and  Siamese Networks in particular. For instance, Mayer et al. \cite {mayer2018learned}, proposed using a Siamese Network to predict whether
pairs of image patches come from the same camera model.
Notably, in \cite{Cozzolino2018NoiseprintAC}, a Siamese Network was employed to extract a camera  model fingerprint, called noiseprint,  from image patches, that can be used for image forgery localization.  In \cite{wang2022eyes} a method  is proposed that utilizes  Siamese Networks to reveal inter-eye symmetries and  inconsistencies for GAN face detection.
Additionally, \cite{Huh_2018_ECCV}, utilized a Siamese network to reveal whether patches from different images have consistent meta-data, facilitating the localization of spliced image content.
To the best of our knowledge, this paper is the first attempt to exploit Siamese Networks for synthetic image attribution.
 
Below, we will briefly present the  state of the art of synthetic  image generation and  image attribution.


	\begin{figure*}[!htb]
		\centering
		\includegraphics[clip,width=0.7\textwidth]{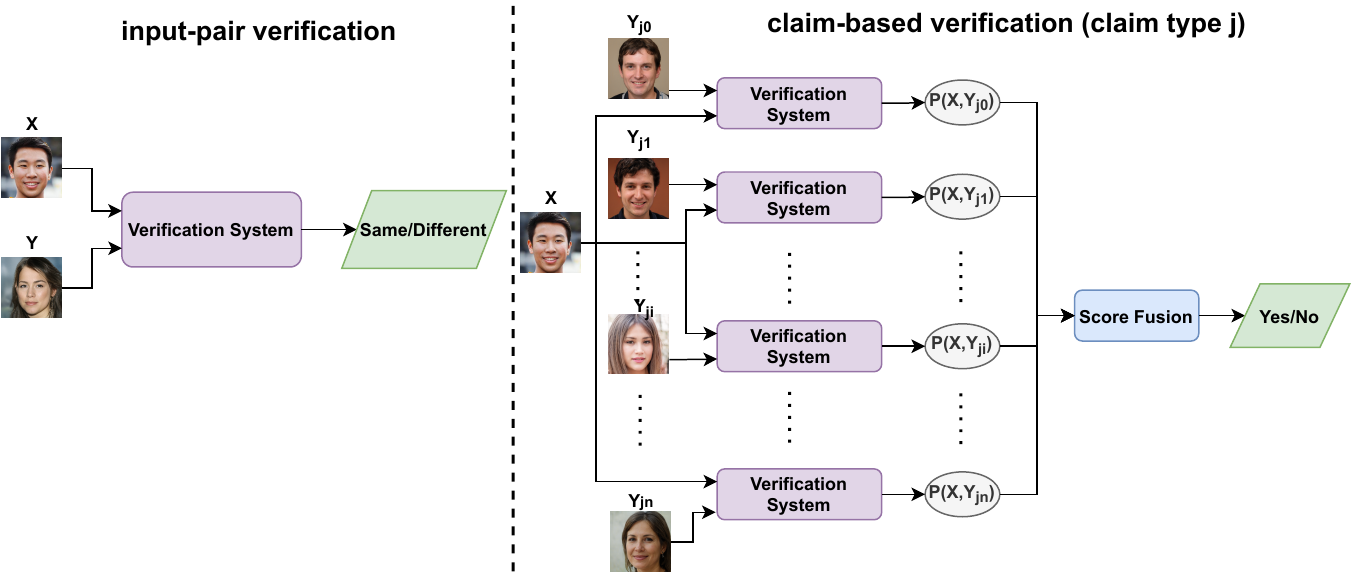}
		\caption{Verification scenarios considered in the paper.}
		\label{fig:framework}
	\end{figure*}
 
	\subsection{Image generation}
  After the early attempts that could only generate low-resolution images, nowadays generative models can produce very high-quality, high-resolution images. The first approaches generating low-resolution images were all based on GAN technology, e.g., BEGAN or BigGAN \cite{berthelot_began_2017,brock_biggan_2019}. An impressive advance was marked by the emergence of the ProGAN architecture \cite{karras_progan_2018}, capable of producing $1024\times1024$ resolution images by using a progressive learning strategy. The quality of the generated high-resolution images was further improved by the StyleGAN series, and in particular, StyleGAN2 \cite{karras_stylegan2_2020,karras_stylegan2ada_2020} and StyleGAN3 \cite{karras_stylegan3_2021}.
  More recently, taming transformers \cite{esser_trans_2021}, a.k.a. VQGANs, combined the power of transformer architectures with the convolutional approach of generative models for image synthesis.
  They utilize encoder-decoder architectures and transformer-based modules to generate high-quality images with coherent image structures. A drawback of GANs is that they suffer from mode collapse, according to which generators tend to produce a small variety of data that is not as diverse as real-world data.
Notable achievements in this direction have been made by diffusion models. Diffusion models smoothly perturb data by adding noise, then reverse this process to generate samples from noisy images (denoising). The pioneering work is the Denoising Diffusion Probabilistic Models (DDPM) \cite{ho_ddpm_2020}, which first demonstrated the model's ability to generate high-quality samples with high levels of detail. Later on, methods applying the diffusion process in the latent space have been proposed (see, for instance, \cite{vahdat_lsgm_2021, rombach_ldiff_2022}).

\subsection{Synthetic image attribution}
\label{ssec.sota.sia}

 The problem of attributing the image to the synthetic model that generated it has been addressed through both active and passive approaches. Active methods involve injecting specific information, e.g., an artificial fingerprint \cite{yu2021artificial}, into the generated images during the generation process, which can later be used to identify the source model.
	On the other hand, passive methods rely on the presence in the generated images of intrinsic artifacts (namely model fingerprints), that are peculiar to the specific model and that can be used to attribute the images to the source model.
Marra et al. \cite{marra2019gans} revealed that each GAN leaves its specific fingerprint in the images it generates. The averaged noise residual image can be taken as a GAN fingerprint. Yue et al. \cite{yu2019attributing} replace the hand-crafted fingerprint formulation in \cite{marra2019gans} with a learning-based one, decoupling the GAN fingerprint into a model fingerprint and an image fingerprint. Frank et al. \cite{frank2020leveraging} and Joslin et al. \cite{joslin2020attributing} perform model attribution considering features in the transformed domain.
	In addition to model-level attribution techniques, researchers have started proposing approaches that address the attribution problem at the architecture-level \cite{yang2021learning,yang2022deepfake,bui2022repmix}, whose goal is to attribute the synthetic images to the source architecture, regardless of how the generative models have been trained, fine-tuned or retrained with a different dataset of pristine images or with different configurations.

 All the above model-level and architecture-level methods work in a closed-set setting, that is when they are asked to analyze images produced by models included in the training set, in the model-level attribution case, or by (possibly unknown) models from known architectures, in the architecture-level attribution case, and fail in the open-set scenario.
 The problem of open-set classification has been addressed in several forensic tasks, like  camera model identification \cite{DEOCOSTA201492,goljan2007identifying}.
With regard to the forensic analysis of synthetically generated contents, while the problem of generalization to unknown  manipulation methods  has been considered recently by several works dealing with synthetic image detection, e.g., \cite{YANG202298,nadimpalli2022improving},
only a few methods have been proposed dealing with synthetic image attribution in open-set scenario,  at model-level
\cite{girish2021towards,yang2023progressive} and at architecture-level \cite{wang2023open}.
These methods rely on the introduction of a rejection option, whereby images generated by models or architectures that were not seen during training are rejected to avoid making a wrong prediction. Such methods, then, must be retrained with samples coming from the new models/architectures if the analysts want to be able to attribute the samples to these architectures/models.
The method in \cite{girish2021towards}, for instance, permits the attribution to any new class, assuming that a few labeled images of such a class are available. These images are exploited to derive the distribution of the new class in the latent space.

As we already stated, in this paper, we adopt a different perspective and treat the open-set architecture attribution problem as a verification task, whereby images produced by out-of-set architectures are handled naturally, without introducing a rejection option.

	\section{Proposed Verification System}
	\label{sec.arch}
	The proposed verification system for synthetic image attribution is illustrated in Fig. \ref{fig:framework}. The following verification scenarios are considered:
	\begin{itemize}
		\item Given two input images $X$ and $Y$, verify whether they are produced by the same generative architecture or not (input pair verification).
		\item Given an input image $X$ and a claim on the generating  architecture, verify whether $X$ has been produced by the claimed architecture or not (claimed-based verification).
	\end{itemize}

	In the first scenario, the system is fed with the input pair $(X, Y)$. The label $m$ associated with the input pair is equal to 0 if $X$ and $Y$ have been generated by the same architecture, $1$ otherwise. By indicating with $\hat{m}$ the output of the system, and with $p(X,Y)$ the probability score, we have $\hat{m} = 0$ if $p(X,Y) < 0.5$, $\hat{m} = 1$ otherwise.
	In the second scenario, the verification works by considering one or multiple reference images $Y_j$ generated from the claimed architecture (of Type-$j$) and evaluating the system with the resulting pairs. In the multiple-reference case, given a dataset of references $D_j$, all the pairs $(X, Y_{ji})$,  $Y_{ji} \in D_j$, are tested and the final decision (Yes/No) is taken by fusing the outputs according to some fusion strategy. In our experiments, we considered both majority voting and score-level fusion. The latter gave the best performance. In particular, the best results were achieved by considering the minimum probability score. The proposed verification framework naturally works in an open-set scenario, where one of the two inputs or both inputs come from an architecture that has not been used for training (with reference to the second verification scenario, either the input $X$, or the claim, or both, may come from an unknown architecture).
		\begin{figure}[!ht]
		\centering
		\includegraphics[clip,width=0.48\textwidth]{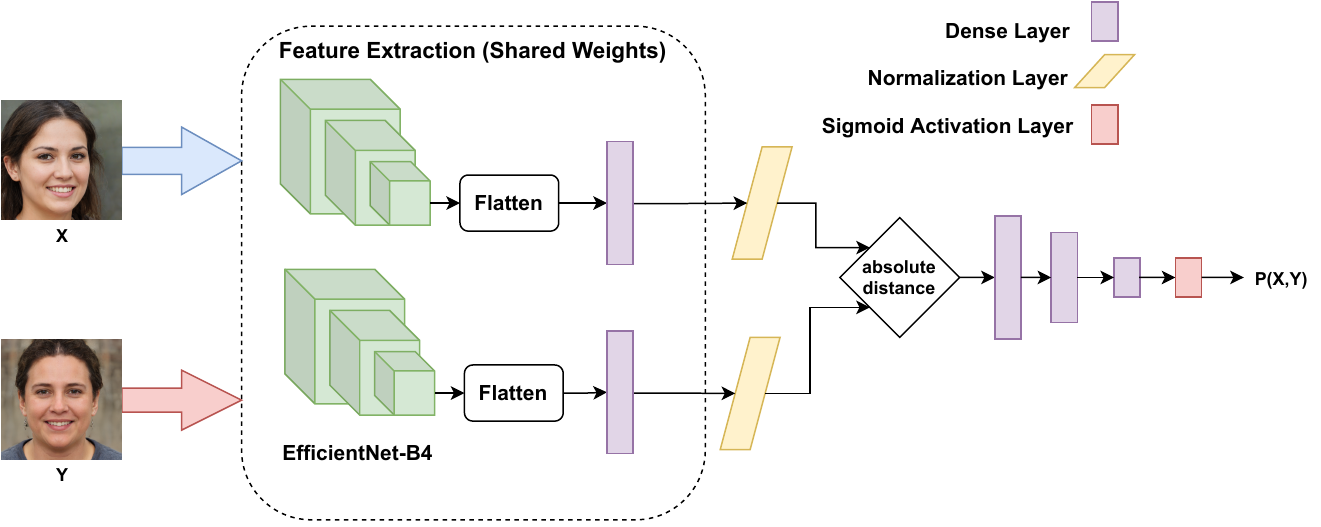}
		\caption{High-Level Architecture for the verification task.}
		\label{fig:Arch}
	\end{figure}

	\subsection{Siamese Network-based architecture}
	\label{ssec.arch}

    The system we are proposing to address the tasks described in Fig. \ref{fig:framework} relies on a Siamese Network (SN) architecture, see Figure \ref{fig:Arch}. It consists of two parts, the feature extraction part, and the decision-making part.
	Feature extraction is performed by an SN that is based on EfficientNet-B4 \cite{tan_efficient_2019} as a backbone for each of the two branches with shared weights. The input image size for each branch is $380\times380$.
 The output of each branch is flattened and then fed as input to a dense layer with size input neurons. The feature embedding, then, consists of 512 elements.
The features are input to a normalization layer, then the point-wise absolute distance between the two output vectors is computed. The distance vector enters the decision-making network, consisting of three consecutive dense layers of sizes 256, 64, and 1 respectively. The final probability scores are obtained by inputting the output of the last dense layer into a sigmoid activation layer.
In our experiments, we also tried other backbone networks to implement the Siamese branches, based on ResNet \cite{he_resnet_2016}
and SWIN transformers \cite{liu_swin_2021}.
While we got perfect results with all these networks in the closed-set setting, the EfficientNet backbone is the one giving the best result in the open-set setting.
	
	\section{Methodology}
	\label{sec.method}

	\subsection{Datasets}
	\label{ssec.method.dataset}

 	\begin{figure}[!htbp]
		\centering
		\includegraphics[clip,width=0.48\textwidth]{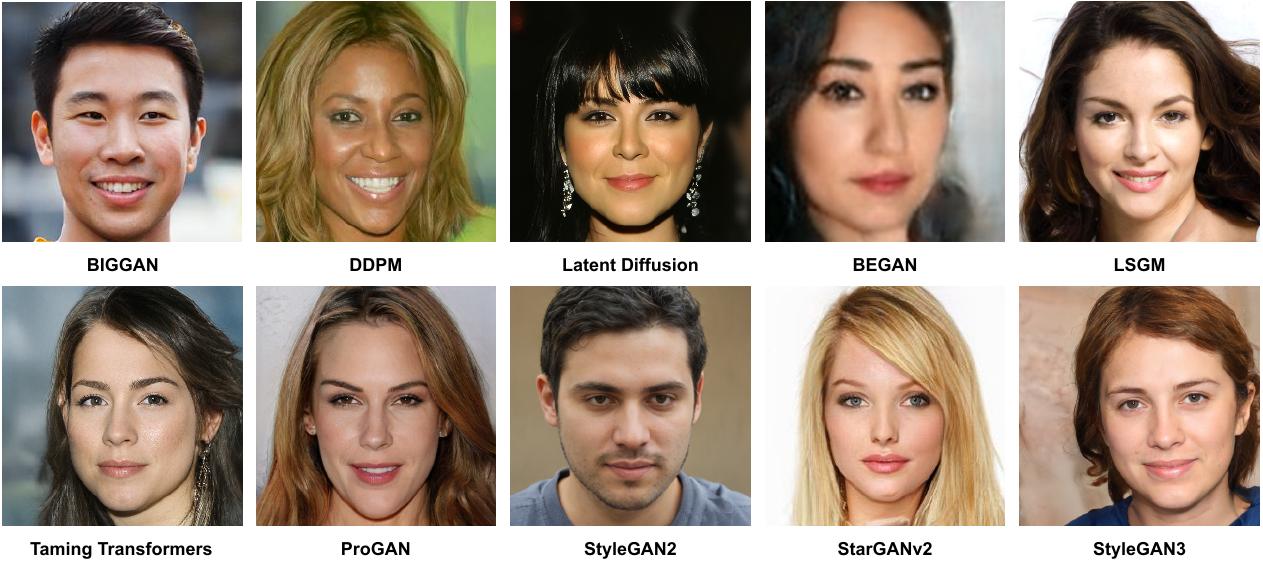}
		\caption{Examples of synthetic images from the 10 architectures.} 
		\label{fig:dataset}
	\end{figure}

	To train and test the verification system, we considered a total of 10 architectures, reported below, including GANs, diffusion models, and transformers. Specifically, we considered the following generative architectures:
	BigGAN \cite{brock_biggan_2019}, BEGAN \cite{berthelot_began_2017}, ProGAN \cite{karras_progan_2018}, StyleGAN2 \cite{karras_stylegan2_2020}, StarGANv2 \cite{choi_starganv2_2020}, StyleGAN3 \cite{karras_stylegan3_2021}, DDPM \cite{ho_ddpm_2020}, Latent Diffusion \cite{rombach_ldiff_2022}, LSGM \cite{vahdat_lsgm_2021} and taming transformers \cite{esser_trans_2021}.

The generative models utilized for the experiments  correspond to  models pre-trained on the FFHQ dataset \cite{karras_ffhq_2019} and the CelebA datasets \cite{liu_faceattributes_2015}, with different configuration parameters, made available in the various repositories.
 In particular, for StyleGAN3, we considered the two best configurations, namely \textit{t} and \textit{r} \cite{karras_stylegan3_2021}, trained on different real datasets and with different resolutions. For StyleGAN2, the best performing configuration, namely configuration \textit{f}, is used for training \cite{karras_stylegan2_2020}.
	
	Starting from this pool of architectures, three different splittings of in-set and out-of-set architectures were considered, with 5 in-set and 5 out-of-set architectures each, named config1, config2, and config3.
	The details of the splittings are reported in Table \ref{tab:data}.
    We observe that in the first and second configurations, a mixture of GANs, diffusion architectures, and transformers were considered as in-set, while in the third configuration, only GANs are included as in-set architectures.
	The in-set architectures are used to train the Siamese verification network, while the out-of-set architectures are only considered for testing. For each in-set architecture, we considered 48,000 images, split into training, validation, and testing sets according to the following numbers 45000:2500:500. For each out-of-set architecture, 500 images were considered for testing.
	Figure \ref{fig:dataset} shows an example of generated images for every selected architecture.
	\begin{table}
		\centering
  \caption{Dataset splitting information. Architectures split (in-set and out-of-set) considered in our experiments.
  }
		\LARGE
		\renewcommand\arraystretch{2}
		\resizebox{\linewidth}{!}{
			\begin{tabular}{c|c|c|c|c}
				\hline
				& Domain & Config1 & Config2 & Config3\\
				\hline
				\multirow{2}{*}{\textbf{In-set}} & FFHQ & \makecell{Latent diffusion,\\ Taming transformers, \\StyleGAN2-f} & \makecell{StyleGAN2-f, \\Latent diffusion} & \makecell{StyleGAN2-f, \\StyleGAN3
    }\\
				\cline{2-5}
				& CelebA & \makecell{Latent diffusion, \\DDPM, BEGAN} & \makecell{BigGAN, ProGAN, \\Latent diffusion, LSGM} & \makecell{ProGAN, BEGAN, \\BigGAN}\\
				\hline
				\multirow{2}{*}{\textbf{Out-of-set}} & FFHQ & StyleGAN3, StarGAN2 & \makecell{StyleGAN3, \\Taming transformers} & \makecell{Latent diffusion,\\ Taming transformers}\\
				\cline{2-5}
				& CelebA & LSGM, ProGAN, BigGAN & \makecell{Taming transformers, \\BEGAN, DDPM, \\StarGAN2} & \makecell{Latent diffusion,\\ Taming transformers, \\LSGM, DDPM, \\StarGAN2}\\
				\hline
		\end{tabular}}
		 \vspace{-0.3cm}
		\label{tab:data}
	\end{table}
	As we said,  to produce the images, we used
	pre-trained models released by the authors in the online repositories.
	\subsection{Siamese Network training}
	\label{ssec.method.train}
	We trained three different SN-based verification models, one for each configuration of in-set and out-of-set architectures, namely config1, config2, and config3.
	The SN-based architecture is trained on a dataset of paired inputs, corresponding to images produced by the same or different architectures, hereafter referred to as positive and negative pairs. For every configuration, the dataset is built from the in-set training dataset as follows: each image is coupled with another image from the same architecture to build a positive pair, and another image is selected randomly from a different architecture to build a negative pair. In this way, the SN is trained on a balanced dataset. Specifically, the training dataset is made up of 45000 $\times$ 5 (no. of images per arch  $\times$  no. of in-set arch) negative pairs and the same number of positive pairs, for a total of 450.000 pair.
	
	In all cases,  training was carried out in two distinct phases: the feature extraction phase and the decision phase. In the first phase,  the two SN branches are trained for 100 epochs, starting from an EfficientNet-B4 model pre-trained on ImageNet, with Adam optimizer and learning rate equal to 0.0001, using the early stopping condition. The network is trained using contrastive loss \cite{hadsell_contrastive_2006}, defined as
		\begin{equation}
		\label{eqn:cLoss}
		\mathcal{L}= (1-m) \cdot d_E^2+ m \cdot [\max(0,h-d_E)]^2,
		\end{equation}
  where $d_E$ is the Euclidean distance between the output of the branches of the SN (embeddings) and $h$ is a  margin hyperparameter that enforces a minimum distance between the two embeddings. We set $h$ to 1 in the experiments.
	The contrastive loss enforces the embeddings of the images in the latent space to be far away whenever the images come from different architectures and close to each other when they belong to the same architecture. Augmentation is performed during training. In particular, we considered JPEG compression, random color transformations (brightness, contrast, saturation, and hue), and random flip.
 The JPEG quality factors are randomly selected within the range [70-100]. For saturation, a random factor between 0.5 and 1 is considered, while hue is adjusted using a random factor between -0.2 and 0.2. Similarly, brightness undergoes modification with a random factor between -0.2 and 0.2, and contrast enhancement is applied with a random factor between 0.2 and 0.5.
 Each type of augmentation is applied to the image with a probability of 0.3. Therefore, different images undergo a different number and different types of augmentations.
	Once the embeddings have been obtained, in the second phase, the weights of the feature extraction network are frozen and the three dense layers following the normalization and the absolute distance layer, responsible for the decision, are trained. The binary cross-entropy (BCE) loss is used for training these layers (decision-making network).
	The layers are trained for 20 epochs with Adam optimizer and a learning rate equal to 0.0001, with an early stopping condition.
The code is made publicly available for reproducibility at the link \url{https://
github.com/lydialy8/openset_attribution_synthetic_images.git.}
 
	\subsection{Testing  procedure}
	\label{ssec.method.test}
	
	We evaluated the proposed method by considering two testing scenarios for verification: one-vs-one and one-vs-many. In the one-vs-one case, each input image in the test set is paired with images generated by the 10 architectures (5 in-set, 5 out-of-set), chosen at random from the test set, thus getting a total of 5000$\times$10 (10\% positive pairs and 90\% negative pairs). Then the SN-based model is evaluated on those pairs. The one-vs-one tests measure the performance of the system in the input-pair verification scenario depicted in Fig.\ref{fig:framework}, and in the claim-based verification scenario, when only one (random) reference is used to verify the claim.
	
	The one-vs-many test setting measures the performance of the system in the claim-based verification scenario when multiple references are available. In our experiments, we considered $100$ reference images. Given a test input image and a claim on the architecture (10 possible claims are considered to correspond to all the architectures) - say Type $j$, we paired each input image with $D_j = 100$ reference images from the claimed architecture. The reference images are randomly selected from the test set.
	The final decision is taken by considering either the mean or the minimum probability score (the latter resulting in the best results). Formally, we consider, respectively, the statistic $(1/|D_j|) \sum_{i \in D_j}  p(X, Y_{ji})$, and  $\min_{i \in D_j} p(X, Y_{ji})$.

	\subsection{Comparison with classification approaches}
	\label{ssec.method.test.class}
	Given that the verification framework proposed in this paper to address the synthetic attribution task is a novel one, no baseline and state-of-the-art methods can be considered for comparison. In order to show the good capabilities of our system in learning good embeddings for the attribution task, we exploited the SN-based verification model inside a classification framework and run a comparison with existing methods for the classification of synthetic attribution in an open-set scenario. In particular, for every configuration of  in-set/out-of-set architectures in Table \ref{tab:data}, a classifier with rejection is obtained as follows: \begin{enumerate*}[label=\textnormal{\roman*)}]
 \item we chose one representative image for every in-set architecture. Specifically, the cluster centroid of the validation sub-dataset (corresponding to the architecture) is considered;
 \item the input image is paired with the 5 representative images obtained at step 1, and the SN-based architecture is tested with these pairs;\item the pair associated with the minimum score is considered.
 \end{enumerate*}
	Formally, let $Z_j$, $j = 1,..,5$ denote the 5 centroids. Given an image $X$, the final classification score associated with $X$ is $\min_{j = 1,\cdots 5} p(X, Z_j)$ and the decision is made for the closed-set architecture $i^*$ that achieving the minimum. Rejection is performed by exploiting the so-called Maximum Softmax Probability (MSP) \cite{hendrycksbaseline}. According to this approach, low confidence in the predicted class reflects the uncertainty of the network prediction, providing evidence that the input sample belongs to an out-of-set class. Then, given a threshold $t$, the output of the classifier $j^*$ is accepted if $p(X, Z_{j^*}) < t$ (lower scores correspond to higher confidences for the 'Same'/'Yes' class in our case), rejected otherwise (unknown input declared).

	\section{Experimental results}
	\label{sec.results}
	
	In this section, we report the performance of the proposed system in the closed and open-set cases and the results of the generalization tests, when unknown models are considered for the same in-set architectures. Finally, we report the comparison results, obtained by considering the classification with rejection system described in Section \ref{ssec.method.test.class}.
	
	\subsection{Verification results}
	\label{ssec.results.verif}
	The results in the one-vs-one setting are reported and discussed below. In Table \ref{table:config1_ref1_closedset_acc}, we report the Accuracy (Acc) of the verification task in the closed-set scenario, when $X$ and $Y$ are produced by in-set architectures, for the 3 configurations. These results show that in the closed-set scenario perfect verification (Acc = 1) can always be achieved by our system. The verification performance in the closed and open-set settings are reported in Table \ref{table:config1_ref1} for each architecture, that is for $Y$ belonging to each of the 10 architectures. The average Area Under Curve  (AUC) of the ROC curve and the probability of correct detection for a false alarm rate equal to 5\% (pd@0.05), are reported for each architecture. The average is computed for the negative pairs over both in-set and out-of-set architectures (9 architectures in total). We observe that the results corresponding to in-set architectures refer to a mixture of closed and open-set scenarios, given that $Y$ may either belong to an in-set or out-of-set architecture (with probability 50\%). Said differently, at least one input of the pair comes from the in-set architectures in this case. Instead, the results reported corresponding to out-of-set architectures refer to the open set scenario, where at least one input of the pair, or both inputs (with probability 50\%) are produced by out-of-set architectures. By looking at this table, we see that when at least one of the two inputs comes from a known architecture, the verification is perfect or almost perfect. In particular, focusing on config1, we see that the AUC is 1 in 4 out of 5 cases (in which the pd@0.05 is also perfect) and 0.94 in the other case.  Similar results are observed in the other configurations.
	The verification performance decreases, still remaining pretty good, in cases where at least one or both inputs come from unknown architectures (results associated with out-of-set architectures). Overall, similar behavior and results are obtained in the three configurations.
	
	\begin{table}[!htbp]
 \centering
 		\caption{Closed-set verification results (Acc).
  }
   \vspace{-0.3cm}
 		\resizebox{0.3\textwidth}{!}{
		\begin{tabular}{c|c|c|c|}
			\cline{2-4}
			\rowcolor[HTML]{EFEFEF}
			\cellcolor[HTML]{FFFFFF}\textbf{}                               & \textbf{Config1}          & \textbf{Config2} & \textbf{Config3} \\ \hline
			\multicolumn{1}{|c|}{\cellcolor[HTML]{EFEFEF}\textbf{Accuracy}} & \cellcolor[HTML]{FFFFFF}1 & 1                & 1                \\ \hline
		\end{tabular}}

		\label{table:config1_ref1_closedset_acc}
	\end{table}

	\begin{table}[!htbp]
 \caption{Verification results (AUC and pd@0.05) in closed and open-set. The cells with green backgrounds indicate in-set architectures, while the white backgrounds indicate out-of-set architectures in the various configurations.}

		 \vspace{-0.3cm}
		\resizebox{0.48\textwidth}{!}{
			\begin{tabular}{|c|cc|cc|cc|}
				\hline
				\rowcolor[HTML]{EFEFEF}
				\cellcolor[HTML]{EFEFEF}                                        & \multicolumn{2}{c|}{\cellcolor[HTML]{EFEFEF}\textbf{Config1}}                            & \multicolumn{2}{c|}{\cellcolor[HTML]{EFEFEF}\textbf{Config2}}                            & \multicolumn{2}{c|}{\cellcolor[HTML]{E7E6E6}\textbf{Config3}}                            \\ \cline{2-7}
				\rowcolor[HTML]{EFEFEF}
				\multirow{-2}{*}{\cellcolor[HTML]{EFEFEF}\textbf{Generating Architecture}} & \multicolumn{1}{c|}{\cellcolor[HTML]{EFEFEF}\textbf{AUC}} & \textbf{pd@0.05}             & \multicolumn{1}{c|}{\cellcolor[HTML]{EFEFEF}\textbf{AUC}} & \textbf{pd@0.05}             & \multicolumn{1}{c|}{\cellcolor[HTML]{EFEFEF}\textbf{AUC}} & \textbf{pd@0.05}             \\ \hline
				\rowcolor[HTML]{E2EFDA}
				\cellcolor[HTML]{EFEFEF}\textbf{Latent Diffusion}               & \multicolumn{1}{c|}{\cellcolor[HTML]{E2EFDA}1}            & 1                            & \multicolumn{1}{c|}{\cellcolor[HTML]{E2EFDA}1}            & 1                            & \multicolumn{1}{c|}{\cellcolor[HTML]{ffffff}0.91}         & \cellcolor[HTML]{ffffff}0.76 \\ \hline
				\rowcolor[HTML]{ffffff}
				\cellcolor[HTML]{EFEFEF}\textbf{DDPM}                           & \multicolumn{1}{c|}{\cellcolor[HTML]{E2EFDA}0.94}         & \cellcolor[HTML]{E2EFDA}0.74 & \multicolumn{1}{c|}{\cellcolor[HTML]{ffffff}0.85}         & 0.68                         & \multicolumn{1}{c|}{\cellcolor[HTML]{ffffff}0.91}         & 0.8                          \\ \hline
				\rowcolor[HTML]{ffffff}
				\cellcolor[HTML]{EFEFEF}\textbf{Taming transformers}            & \multicolumn{1}{c|}{\cellcolor[HTML]{E2EFDA}1}            & \cellcolor[HTML]{E2EFDA}1    & \multicolumn{1}{c|}{\cellcolor[HTML]{ffffff}0.88}         & 0.72                         & \multicolumn{1}{c|}{\cellcolor[HTML]{ffffff}0.84}         & 0.66                         \\ \hline
				\rowcolor[HTML]{E2EFDA}
				\cellcolor[HTML]{EFEFEF}\textbf{StyleGAN2}                      & \multicolumn{1}{c|}{\cellcolor[HTML]{E2EFDA}1}            & 1                            & \multicolumn{1}{c|}{\cellcolor[HTML]{E2EFDA}1}            & 1                            & \multicolumn{1}{c|}{\cellcolor[HTML]{E2EFDA}1}            & 1                            \\ \hline
				\rowcolor[HTML]{E2EFDA}
				\cellcolor[HTML]{EFEFEF}\textbf{BEGAN}                          & \multicolumn{1}{c|}{\cellcolor[HTML]{E2EFDA}1}            & 0.99                         & \multicolumn{1}{c|}{\cellcolor[HTML]{ffffff}0.95}         & \cellcolor[HTML]{ffffff}0.79 & \multicolumn{1}{c|}{\cellcolor[HTML]{E2EFDA}1}            & 1                            \\ \hline
				\rowcolor[HTML]{ffffff}
				\cellcolor[HTML]{EFEFEF}\textbf{StyleGAN3}                      & \multicolumn{1}{c|}{\cellcolor[HTML]{ffffff}0.9}          & 0.81                         & \multicolumn{1}{c|}{\cellcolor[HTML]{ffffff}0.9}          & 0.81                         & \multicolumn{1}{c|}{\cellcolor[HTML]{E2EFDA}0.97}         & \cellcolor[HTML]{E2EFDA}0.92 \\ \hline
				\rowcolor[HTML]{ffffff}
				\cellcolor[HTML]{EFEFEF}\textbf{LSGM}                           & \multicolumn{1}{c|}{\cellcolor[HTML]{ffffff}0.84}         & 0.68                         & \multicolumn{1}{c|}{\cellcolor[HTML]{E2EFDA}1}            & \cellcolor[HTML]{E2EFDA}0.98 & \multicolumn{1}{c|}{\cellcolor[HTML]{ffffff}0.7}          & 0.34                         \\ \hline
				\rowcolor[HTML]{ffffff}
				\cellcolor[HTML]{EFEFEF}\textbf{StarGAN v2}                     & \multicolumn{1}{c|}{\cellcolor[HTML]{ffffff}0.88}         & 0.72                         & \multicolumn{1}{c|}{\cellcolor[HTML]{ffffff}0.84}         & 0.68                         & \multicolumn{1}{c|}{\cellcolor[HTML]{ffffff}0.89}         & 0.8                          \\ \hline
				\rowcolor[HTML]{E2EFDA}
				\cellcolor[HTML]{EFEFEF}\textbf{BIGGAN}                         & \multicolumn{1}{c|}{\cellcolor[HTML]{ffffff}0.95}         & \cellcolor[HTML]{ffffff}0.79 & \multicolumn{1}{c|}{\cellcolor[HTML]{E2EFDA}0.9}          & 0.74                         & \multicolumn{1}{c|}{\cellcolor[HTML]{E2EFDA}0.92}         & 0.79                         \\ \hline
				\rowcolor[HTML]{E2EFDA}
				\cellcolor[HTML]{EFEFEF}\textbf{PROGAN}                         & \multicolumn{1}{c|}{\cellcolor[HTML]{ffffff}0.85}         & \cellcolor[HTML]{ffffff}0.68 & \multicolumn{1}{c|}{\cellcolor[HTML]{E2EFDA}1}            & 1                            & \multicolumn{1}{c|}{\cellcolor[HTML]{E2EFDA}1}            & 1                            \\ \hline
		\end{tabular}}
		\label{table:config1_ref1}
	\end{table}

	In Table \ref{table:overall_ref1}, we report the average results for all configurations. The total AUC is averaged over all the possible pairs of inputs, hence considering all the pairs' combinations (in-set vs in-set, in-set vs out-of-set, out-of-set vs in-set, and out-of-set vs out-of-set).
	The open-set AUC instead is computed by considering only the out-of-set vs out-of-set pairs (fully open set), while the closed-set AUC is computed  over the in-set vs in-set pairs.
	The results show that config1 shows better results in the open-set scenario. We observe that in this configuration, the out-of-set set contains (mostly) GAN architectures and a diffusion-type architecture (LSGM), that are also present in the in-set. This is not the case in the other configurations where, for instance, transformers in config2 and both diffusion models and transformers in config3 are only considered as out-of-set, without any of these types of architectures included in the in-set set.

	\begin{table}[!htbp]
       \centering
		\caption{Total, open-set and closed-set AUCs.}
		\resizebox{0.35\textwidth}{!}{
			\begin{tabular}{c|
					>{\columncolor[HTML]{FFFFFF}}c |
					>{\columncolor[HTML]{FFFFFF}}c |
					>{\columncolor[HTML]{FFFFFF}}c |}
				\cline{2-4}
				& \cellcolor[HTML]{EFEFEF}\textbf{Config1} & \cellcolor[HTML]{EFEFEF}\textbf{Config2} & \cellcolor[HTML]{EFEFEF}\textbf{Config3} \\ \hline
				\multicolumn{1}{|c|}{\cellcolor[HTML]{EFEFEF}\textbf{Total AUC}}   & 0.95                                     & 0.93                                     & 0.93                                     \\ \hline
				\multicolumn{1}{|c|}{\cellcolor[HTML]{EFEFEF}\textbf{Open-set AUC}}   & 0.92                                     & 0.81                                     & 0.85                                     \\ \hline
				
				\hline
				\multicolumn{1}{|c|}{\cellcolor[HTML]{EFEFEF}\textbf{Closed-set AUC}}   & 1                                     & 1                                     & 1                                     \\ \hline
		\end{tabular}}
		\label{table:overall_ref1}
	\end{table}
	
	In Table \ref{table:overall_ref100}, we report the average results of the tests one-vs-many, for all the configurations. In all the cases, a slight improvement is observed when the minimum score is considered, compared to the case of one reference only, while the mean score case only improves in a few cases. These results show that using multiple random references for the verification improves the results only slightly. A possible reason is that all the feature vectors for a given architecture tend to cluster close to each other, yielding a similar verification result.

	\begin{table}[!htbp]
 		\caption{Total, open-set, and closed-set AUCs in the one-vs-many setting.
  } \vspace{-0.1cm}
		\resizebox{0.4\textwidth}{!}{
			\begin{tabular}{c|
					>{\columncolor[HTML]{FFFFFF}}c
					>{\columncolor[HTML]{FFFFFF}}c |
					>{\columncolor[HTML]{FFFFFF}}c
					>{\columncolor[HTML]{FFFFFF}}c |
					>{\columncolor[HTML]{FFFFFF}}c
					>{\columncolor[HTML]{FFFFFF}}c |}
				\cline{2-7}
				\multicolumn{1}{l|}{}                                                & \multicolumn{2}{l|}{\cellcolor[HTML]{EFEFEF}\textbf{Config1}}                                     & \multicolumn{2}{l|}{\cellcolor[HTML]{EFEFEF}\textbf{Config2}}                                     & \multicolumn{2}{l|}{\cellcolor[HTML]{EFEFEF}\textbf{Config3}}                                     \\ \cline{2-7}
				\textbf{}                                                            & \multicolumn{1}{c|}{\cellcolor[HTML]{EFEFEF}\textbf{Mean}} & \cellcolor[HTML]{EFEFEF}\textbf{Min} & \multicolumn{1}{c|}{\cellcolor[HTML]{EFEFEF}\textbf{Mean}} & \cellcolor[HTML]{EFEFEF}\textbf{Min} & \multicolumn{1}{c|}{\cellcolor[HTML]{EFEFEF}\textbf{Mean}} & \cellcolor[HTML]{EFEFEF}\textbf{Min} \\ \hline
				\multicolumn{1}{|c|}{\cellcolor[HTML]{EFEFEF}\textbf{Total AUC}}   & \multicolumn{1}{c|}{\cellcolor[HTML]{FFFFFF}0.95}          & 0.96                                 & \multicolumn{1}{c|}{\cellcolor[HTML]{FFFFFF}0.94}          & 0.94                                 & \multicolumn{1}{c|}{\cellcolor[HTML]{FFFFFF}0.94}          & 0.94                                 \\ \hline
				\multicolumn{1}{|c|}{\cellcolor[HTML]{EFEFEF}\textbf{Open-set AUC}}   & \multicolumn{1}{c|}{\cellcolor[HTML]{FFFFFF}0.92}          & 0.94                                 & \multicolumn{1}{c|}{\cellcolor[HTML]{FFFFFF}0.78}          & 0.84                                 & \multicolumn{1}{c|}{\cellcolor[HTML]{FFFFFF}0.87}          & 0.85                                 \\ \hline
				
				\multicolumn{1}{|c|}{\cellcolor[HTML]{EFEFEF}\textbf{Closed-set AUC}}   & \multicolumn{1}{c|}{\cellcolor[HTML]{FFFFFF}1}          & 1                                & \multicolumn{1}{c|}{\cellcolor[HTML]{FFFFFF}1}          & 1                                 & \multicolumn{1}{c|}{\cellcolor[HTML]{FFFFFF}1}          & 1                                 \\ \hline
		\end{tabular}}
		\label{table:overall_ref100}
	\end{table}
	
	\subsection{Generalization tests}
	\label{ssec.results.gen}

We also ran some generalization tests in order to prove the capability of the proposed system to perform attribution when {\em unknown} models are considered for the various in-set architectures during testing. The purpose of these tests is to show that the system works as supposed to and indeed attributes images to the architecture and not to the single models. 
In all generalization tests, positive pairs were formed by pairing images from the unknown models with random images from the known models from the same in-set architecture, while negative pairs were created by pairing images from unknown models with random images from different in-set architectures.
 
 	In particular, we considered new models from the in-set architectures that were trained:  i) on a different dataset; ii) by using a different training methodology; and iii) using different training configurations.
 For case i), we tested a system trained in config1, considering as unknown model  a taming transformer model trained on the  CelebA dataset (the models considered during training for the taming transformer architectures are trained on the FFHQ dataset). Therefore, in this case, positive pairs are formed by generating images from taming transformers trained respectively on  FFHQ and CelebA. For the negative pairs, images from the taming transformer model trained CelebA (namely, the unknown) are coupled with images generated from known models for different architectures. 
 In case ii), we tested the systems trained in all configurations considering as  unknown model the  StyleGAN2-ada \cite{karras_stylegan2ada_2020} (the systems are trained with images from StyleGAN2-f), which employs an adaptive discriminator augmentation mechanism for training stability in limited data regimes. Finally, for case iii), we evaluated the system trained in config3 using, as unknown model, a StyleGAN model obtained through retraining on unaligned ({\em -u-}) high-resolution faces (FFHQ-U) with resolution 1024$\times$1024.

The results reported in Table \ref{table:generalization} demonstrate that the system generalizes well in all scenarios, achieving an AUC and Accuracy equal to 1.


	\begin{table}[!htbp]
   		\caption{Results with models trained with different datasets, parameters, and training procedure (in the last line,   the number in the models' names refers to the image resolution).
     }
    \vspace{-0.3cm}
		\resizebox{0.48\textwidth}{!}{
			\begin{tabular}{|
					>{\columncolor[HTML]{EFEFEF}}c |c|c|c|c|c|}
				\hline\textbf{Architecture} & \cellcolor[HTML]{EFEFEF}\textbf{\makecell{Type of \\Mismatch}} & \cellcolor[HTML]{EFEFEF}\textbf{Train} & \cellcolor[HTML]{EFEFEF}\textbf{Test } & \cellcolor[HTML]{EFEFEF}\textbf{AUC} & \cellcolor[HTML]{EFEFEF}\textbf{ACC} \\ \hline
				\textbf{\makecell{Taming Transf\\Config1}}   &  Dataset                                 & FFHQ                                   & CelebA           & 1      & 1                 \\ \hline\textbf{\makecell{StyleGAN2\\Config1}}     &  \makecell{Training\\Methodology}                                 & StyleGAN2-f                              & StyleGAN2-ada    & 1 & 1                      \\\hline \textbf{\makecell{StyleGAN2\\Config2}}               &  \makecell{Training\\Methodology}                   & StyleGAN2-f                              & StyleGAN2-ada   & 1     & 1                   \\\hline  
 \textbf{\makecell{StyleGAN2\\Config3}}            &  \makecell{Training\\Methodology}                            & StyleGAN2-f                              & StyleGAN2-ada     & 1  & 1                   \\ \hline
 \textbf{\makecell{StyleGAN3\\Config3}}                                & Configuration      &  \makecell{StyleGAN3-\\(t-1024/t-u256/r)}                                 & \makecell{StyleGAN3-\\t-u1024}  & 1 & 1 \\\hline
		\end{tabular}}

		\label{table:generalization}
	\end{table}
	
	\subsection{Comparison results}
	\label{ssec.results.sota}
	In this section, we report the results of the experiments that we run considering a classifier built by starting from the proposed verifier, as detailed in Section \ref{ssec.method.test.class}. This system is compared with state-of-the-art classification methods, namely the  PCSSR and RCSSR variants of the method in \cite{huang2022class}, recently proposed in the general literature of machine learning for  open-set classification, and the ResVit method \cite{wang2023open} for the classification of synthetic manipulation and attribution in open-set settings. All these methods perform classification with a rejection option, thus rejecting unknown samples.
	As to the performance metrics, following \cite{wang2023open}, we use the Accuracy  to measure the closed-set performance, and the AUC to  measure the rejection performance. The results reported in Table \ref{table:sota} show that the proposed classifier is the one obtaining the best average performance in all the three configurations of in-set and out-of-set architectures, with a perfect Acc and an AUC gain which is about 8\% on the average over PCSSR and RCSSR, and 11\% over  \cite{wang2023open}.
	These results show the superior capability of our method based on similarity learning of getting characteristic embeddings for the various architectures.
	Once again, we stress that we considered this framework only for comparison purposes. Indeed, the capabilities of the verification system that we proposed in this paper in the open-set scenario are not limited to sample rejection, given that our system can provide the same functionality in both closed and open-set scenarios.

	\begin{table}[!htbp]
   		\caption{Comparison  of closed-set (Acc) and open-set (AUC) performance of the classifier based on our SN-based model with state-of-the-art classifiers.}
		\resizebox{0.48\textwidth}{!}{
			\begin{tabular}{
					>{\columncolor[HTML]{EFEFEF}}c l|c|c|c|c|}
				\cline{3-6}
				\cellcolor[HTML]{FFFFFF}{\color[HTML]{FFFFFF} \textbf{}}                         &                   & \cellcolor[HTML]{EFEFEF}{\color[HTML]{000000} \textbf{ResVit}\cite{wang2023open}} & \cellcolor[HTML]{EFEFEF}{\color[HTML]{000000} \textbf{PCSSR}\cite{huang2022class}} & \cellcolor[HTML]{EFEFEF}{\color[HTML]{000000} \textbf{RCSSR}\cite{huang2022class}} & \cellcolor[HTML]{EFEFEF}{\color[HTML]{000000} \textbf{Ours}} \\ \hline
				\multicolumn{1}{|c|}{\cellcolor[HTML]{EFEFEF}}                                   & \textbf{Accuracy} & 0.99                                                         & 0.99                                                          & 0.99                                                          & \textbf{1}                                                   \\ \cline{2-6}
				\multicolumn{1}{|c|}{\multirow{-2}{*}{\cellcolor[HTML]{EFEFEF}\textbf{Config1}}} & \textbf{AUC}      & 0.79                                                         & \textbf{0.84}                                                 & 0.83                                                          & 0.82                                                         \\ \hline
				\multicolumn{1}{|c|}{\cellcolor[HTML]{EFEFEF}}                                   & \textbf{Accuracy} & 0.99                                                         & 0.99                                                          & 0.99                                                          & \textbf{1}                                                   \\ \cline{2-6}
				\multicolumn{1}{|c|}{\multirow{-2}{*}{\cellcolor[HTML]{EFEFEF}\textbf{Config2}}} & \textbf{AUC}      & 0.76                                                         & 0.74                                                          & 0.75                                                          & \textbf{0.82}                                                \\ \hline
				\multicolumn{1}{|c|}{\cellcolor[HTML]{EFEFEF}}                                   & \textbf{Accuracy} & 0.99                                                         & 0.99                                                          & 0.99                                                          & \textbf{1}                                                   \\ \cline{2-6}
				\multicolumn{1}{|c|}{\multirow{-2}{*}{\cellcolor[HTML]{EFEFEF}\textbf{Config3}}} & \textbf{AUC}      & 0.68                                                         & 0.66                                                          & 0.64                                                          & \textbf{0.83}                                                \\ \hline
		\end{tabular}}

  \vspace{-0.3cm}
		\label{table:sota}
	\end{table}

	\section{Conclusion}
	\label{sec.conclusion}
We have proposed a novel verification framework to address the problem of synthetic architecture attribution in open set conditions. The experiments we ran demonstrated good performance of our system in both closed and open-set settings when different mixtures of generative architectures of synthetic face images are considered as in-set and out-of-set. Generalization performance is also good, with unknown models correctly attributed to the source architectures. We also showed that when the SN-based verification model is used to build a classifier with a rejection class, the results, we got are superior to those achieved by state-of-the-art methods. Future work will focus on improving the results obtained for the one-vs-many case, by optimizing the choice of the images used as references. Moreover, we will consider synthetic images with different semantic content (beyond the face domain) and investigate the generalization capabilities of the proposed system, when the architectures are trained on other domains to get models producing images belonging to different categories.

\section{Acknowledgements}
 This work has been partially supported by the China Scholarship Council (CSC), file No. 202008370186, the Italian Ministry of University and Research, PREMIER project, under contract PRIN 2017 2017Z595XS-001, FOSTERER project under contract PRIN 2022 202289RHHP, and the Defense Advanced Research Projects Agency (DARPA) and the Air Force Research Laboratory (AFRL) under agreement number FA8750-20-2-1004.
	The U.S. Government is authorized to reproduce and distribute reprints for Governmental purposes notwithstanding any copyright notation thereon.
	\bibliographystyle{model2-names}
	\bibliography{egbib}
\end{document}